\documentclass[manuscript,screen]{acmart}
\usepackage{amsfonts} 
\newcommand{\bbR}{\mathbb{R}}
\newcommand{\bbC}{\mathbb{C}}
\AtBeginDocument{%
  }

\copyrightyear{2026}
\acmYear{2026}
\setcopyright{usgov}
\acmConference[ICONS 2026]{International Conference on Neuromorphic Systems}{August 04--06, 2026}{Chicago, IL, USA}
\acmBooktitle{International Conference on Neuromorphic Systems (ICONS 2026), August 04--06, 2026, Chicago, IL, USA}
\acmDOI{10.1145/3822454.3822484}
\acmISBN{979-8-4007-2808-2/2026/08}

\begin{document}

\title{Phase State Space Models: Parallel, Surrogate-Free Training of Spiking Networks}

\author{Wilkie Olin-Ammentorp}
\email{wolinammentorp@anl.gov}
\orcid{0000-0002-2472-9862}
\affiliation{%
  \institution{Argonne National Laboratory}
  \city{Lemont}
  \state{Illinois}
  \country{USA}
}

\renewcommand{\shortauthors}{Olin-Ammentorp}

\begin{abstract}
    State-space models (SSMs) provide a powerful theoretical framework to enable parallel training of recurrent networks. We expand on previous work adapting SSMs to spiking models to provide a novel interpretation of resonate-and-fire (R\&F) neural networks which is compatible both with real and spiking inputs, parallel and recurrent execution, has clear connections to hyperdimensional (HD) computing, and maintains biologically-realistic features. We demonstrate an implementation of this approach which integrates an STFT, recurrent memory, and attentional features within a single spike-compatible network.
\end{abstract}



\received{29 May 2026}
\received[accepted]{20 June 2026}

\maketitle

\section{Introduction}
State-space models (SSMs) provide a powerful mathematical framework to analyze networks with recurrent features. This framework provides three equivalent views on these networks as continuous, discrete, and convolutional. This equivalence enables powerful parallel training of these networks in the convolutional mode and efficient execution in the discrete and/or continuous mode \cite{guCombiningRecurrentConvolutional}. Previous work has demonstrated that the SSM approach can be successfully adapted to spiking neural networks based on both the integrate-and-fire (I\&F) and resonate-and-fire (R\&F) neuron models \cite{duSpikingStructuredState2024, balPSpikeSSMHarnessingProbabilistic2025, huberScalingResonateandFireNetworks2025}. In this work, we focus on networks based on the R\&F neurons for their biological realism, connection to hyperdimensional (HD) computing, innate recurrent memory, rich repertoire of hardware implementations, and robust mathematical framework. We will expand on each of these topics through the course of this paper, beginning with a novel derivation of a Phase State Space Model ($\varphi$-SSM) based on the R\&F neuron.

\section{Phase State Space Models}
\subsection{Resonate-and-Fire Neurons}
The R\&F neuron can be thought of as a leaky I\&F neuron which has been redefined in the complex plane \cite{izhikevichResonateandfireNeurons2001}. Both can be implemented with the differential update:
\begin{equation}
\label{eqn_rnf}
    dU/dt = k \cdot U(t) + w \cdot I(t)
\end{equation}
where $U$ is the neural potential, $k$ is the leakage, $w$ are the input weights, and $I(t)$ is the input current as a function of time. In an I\&F neuron, these quantities are all real-valued. R\&F neurons make the change of placing potential on the complex plane with $k,U,w \in \bbC$, with currents generally remaining real ($I(t) \in \bbR$). Furthermore, $k$ can be defined as:
\begin{equation}
\label{eqn_leakage}
    k = \lambda + i \cdot \omega
\end{equation}
where parameter $\lambda < 0$ controls the R\&F neuron's ``leakage'' or damping, and $\omega$ controls its angular frequency -- the ``speed'' with which a potential resonates from current to voltage. This simple change gives rise to a bounty of interesting and biologically-relevant behaviors which are absent from the I\&F model, such as bursting and resonance \cite{izhikevichResonateandfireNeurons2001}. Additionally, we can view the complex potential $U(t)$ as a two-dimensional polar value. Throughout, we write $\angle(z) \equiv \arg(z)/\pi \in [-1, 1]$ for the phase angle expressed in half-turns (units of $\pi$):

\begin{equation}
\label{eqn_polar}
  U(t) = r(t)\,e^{i \pi \theta(t)},
  \qquad r(t) = |U(t)|,
  \quad \theta(t) = \angle(U(t)).
\end{equation}
An R\&F neuron ``fires'' when $r(t) > \vartheta$ where $\vartheta$ is the neuron's firing threshold, and its argument or ``angle'' $\theta$ passes through $0$. This spiking transformation is defined as follows:

\begin{equation}
\label{eqn_spiking}
    S(u)=\mathbb{1}\!\left[\,|u| > \vartheta\,\right] \cdot \delta(\theta(u))
\end{equation}
where $\delta$ is the Dirac delta function.
A spike produced by an R\&F neuron defined via these equations does not give us the full information of its sender's complex potential, but it does allow us to infer its phase by communicating when it passes through $0^{\circ}$; in this manner, we can interpret spikes as sparsely encoding and efficiently communicating the instantaneous phases of R\&F neurons.

\subsection{State Space Models}

Previous work has already demonstrated that the SSM model can be successfully extended to R\&F neurons \cite{huberScalingResonateandFireNetworks2025}. However, we provide an alternate formulation which explicitly interprets the spikes of R\&F neurons as communicating phases. As a result, the computational methods derived in this work provide a different set of capabilities and trade-offs.

Briefly, we recapitulate the fundamentals of SSMs. The differential transition of a potential $U$ may be defined as:
\begin{equation}
\label{eqn_ssm}
    dU / dt = A \cdot U(t) + B \cdot X(t)
\end{equation}
where $U$ is a tensor of neural potentials, $A$ is a transition matrix which defines the system's recurrent behavior, $X$ is an external input, and $B$ is a matrix which projects these inputs into the space of $U$. As has been previously noted, the update equation for I\&F and R\&F neurons (Eqn. \ref{eqn_rnf}) is identical to an SSM: if the tensors of the update equation are complex-valued and the external input $X(t)$ represents input currents, we produce the differential update for the membrane potential of a layer of R\&F neurons.
\begin{equation}
\label{eqn_rnf_ssm}
    dU / dt = A \cdot U(t) + B \cdot I(t)
\end{equation}
In some SSM models, the matrix $A$ is non-diagonal, allowing elements of $U$ to directly be influenced by one another. However, this requires neurons with direct access to the internal  potential of other neurons; this direct connection is not biologically realistic, as cellular membranes isolate this information between neurons. As a result, we impose a diagonal structure on $A$. Diagonalization of the matrix $A$ has already been introduced in many SSMs, but is explicitly required in our model in order to maintain locality of information \cite{guCombiningRecurrentConvolutional}. Additionally, the variable $U$ is not directly observed as an output, but is transformed by a second equation:
\begin{equation}
    \label{eqn_ssm2}
    Y (t) = C \cdot U(t) + D \cdot X(t)
\end{equation}
This linear transformation allows the neural states $U$ to be projected into the output space and allows for a residual ``skip'' connection between inputs and outputs.

To adapt this output transformation to R\&F neurons, we must consider that the output of these neurons is not the full complex potential, but an instantaneous phase. Thus, we may reformulate Eqn. \ref{eqn_ssm2} as:
\begin{equation}
    \label{eqn_pssm1}
    Y(t) = S( U(t))
\end{equation}
by substituting an identity matrix for $C$, setting the skip connection $D$ to zero, and applying the spiking transformation $S$ (Eqn. \ref{eqn_spiking}) to $U(t)$. This defines a spiking state-space model on the R\&F neuron model. This model inherently is a non-linear transformation via $S$, and can be chained arbitrarily to form neural networks. The spiking output of Eqn. \ref{eqn_pssm1} can be directly used to form the input signal $I(t)$ for a successive layer.

\subsection{Surrogate-Free Derivatives of Relative Phases}

While Eqn. \ref{eqn_pssm1} allows for a fully spiking, continuous-time system to be solved, we will encounter practical difficulties in executing and optimizing such a system. Namely, truly instantaneous spikes cannot be physically realized, and defining gradients through these jumps is difficult. Usually, this challenge is side-stepped by utilizing a relaxed kernel with a Gaussian or exponential shape, and utilizing this ``surrogate'' to define smooth gradients which can be optimized via backpropagation through time (BPTT) \cite{neftciSurrogateGradientLearning2019}. However, this approach carries its own set of challenges. Computational steps will not provide gradient information until a neuron fires, optimizing the shape and duration of the surrogate kernel is required, and recurrent inference forward in time may require fine-grained temporal steps. We take an alternative approach to allow for the realization of networks which provide gradients even for `silent' neurons, enables fixed-size temporal steps, and does not require surrogate kernels.

Building on our previous work, we note that by fixing the resonant frequency $\omega$ of layers of R\&F neurons --- similarly to how neurons are synchronized by organized, traveling waves in the brain --- a new invariant property is introduced to groups of neurons. By fixing $\omega$, the differences in instantaneous phases between neurons (\textit{relative phase}) is constant through time \cite{olin-ammentorpHyperdimensionalComputingProvides2023}. Thus, at any point in time, the relative phase between two neurons can be decoded:
\begin{equation}
\label{eqn_relphase}
    \theta_{\text{rel}}(u, \omega, t) = \angle\bigl(u \cdot e^{-i\omega t}\bigr)
\end{equation}
where $e^{-i \omega t}$ is a `reference' oscillator whose complex potential represents the phase $0$ at all times. Utilizing this reference potential, the relative motion of another potential $U$ through the complex plane with respect to time is removed, allowing the relative phase to be decoded. This allows us to define an alternative formulation of the previous SSM:
\begin{equation}
    \label{eqn_pssm2}
    Y_{\theta}(t) = \theta_{\text{rel}}( U(t), \omega, t)
\end{equation}
This adjustment to Eqn. \ref{eqn_pssm1} provides a real-valued, angular output of the system at \textit{every} moment in time, rather than only when a neuron spikes. Furthermore, this output is locally continuous and can provide gradients solved by standard automated differentiation (AD) \cite{baydin2018automatic}. No surrogate spiking kernel is required, as we directly solve through the potential of the neuron at each point in time. Furthermore, even neurons below the spiking threshold $\vartheta$ will still contain an angular value which can provide gradient information. Finally, the solution of $Y_{\theta}(t)$ can be calculated at discrete steps which are naturally defined by the resonant frequency $\omega$, as we demonstrate next.

\subsection{Parallel Training, Recurrent Execution}

\begin{figure*}
    \centering
    \includegraphics[width=1\textwidth]{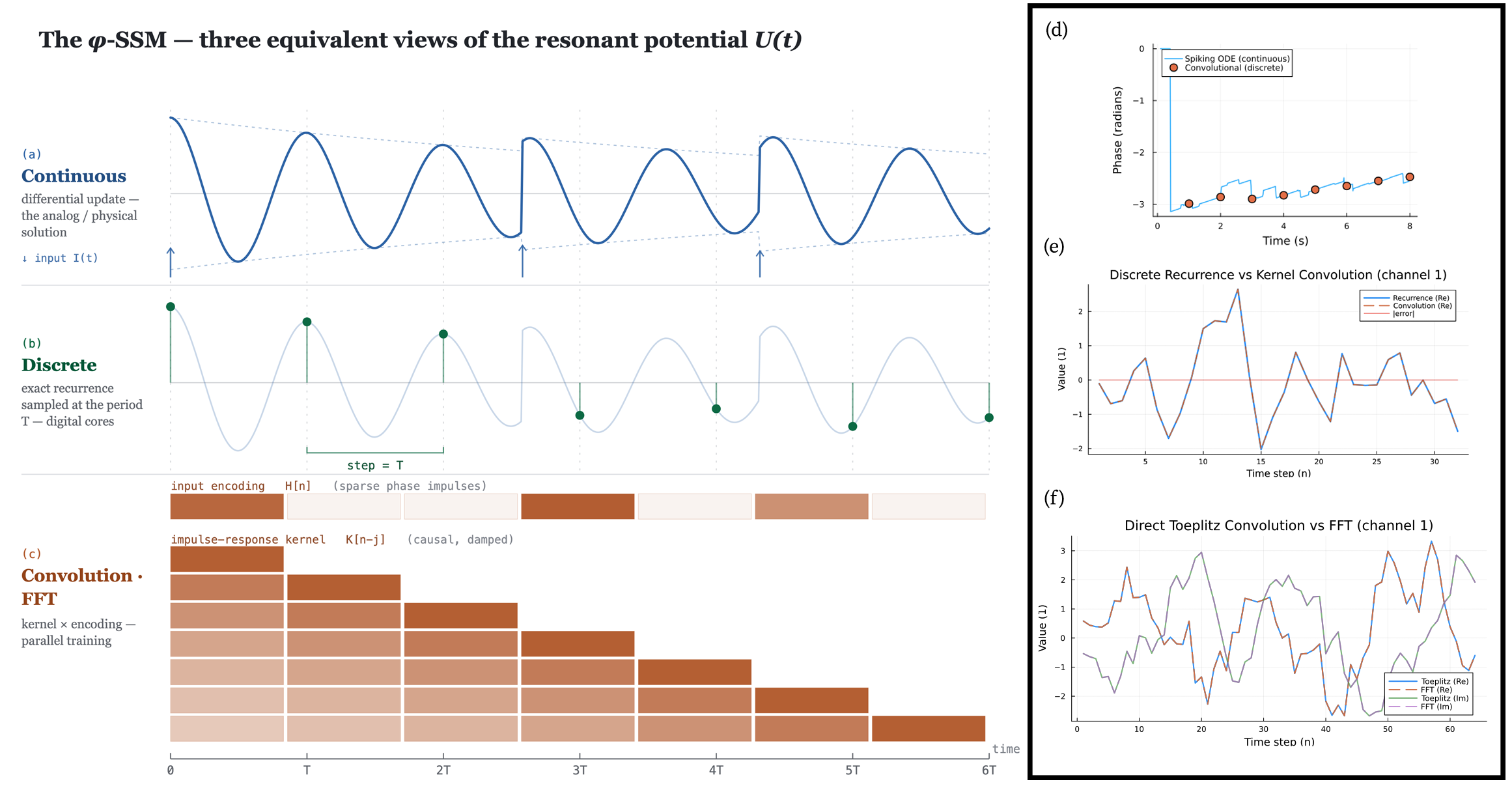}
  \caption{The $\varphi$-SSM provides three equivalent views which can be used to calculate the neural potentials of R\&F neurons through time. The continuous, differential view (a) can be implemented directly via physical oscillators or simulated via differential solvers. The discrete view (b) calculates the same updates at regularly-spaced intervals at multiples of $T$, the resonant period of the neural layer, and can be implemented on digital cores. The convolutional/FFT view (c) requires the storage of all timesteps previously solved iteratively, but offers the advantage of computing the state of all neurons at all timesteps in parallel, and can be implemented on large parallel systems (e.g. GPUs) for accelerated training. All three views produce the same outputs within numerical precision, with comparisons shown in the right sidebox (d-f).}  
  \label{fig:equivalence}
\end{figure*}

We have defined continuous-time differential updates which can be solved to find the neuron potentials (Eqn. \ref{eqn_rnf_ssm}), spiking outputs (Eqn. \ref{eqn_pssm1}), and relative phases (Eqn. \ref{eqn_pssm2}) through time. A differential solver may be applied to these equations to find a solution (Fig. \ref{fig:equivalence}a), but one of the key advantages of an SSM is its ability to be parallelized in execution --- the matrices $A$, $B$, $C$, and $D$ may be reformulated to allow for a solution to the equations at any point in time using a convolutional kernel or fast Fourier transform (FFT). For a real-valued input $I(t)$, standard derivations of the adjusted kernels $\bar{A}$, $\bar{B}$, $\bar{C}$, and $\bar{D}$ using a zero-order hold (ZOH) or bilinear adaptations may be adopted \cite{guCombiningRecurrentConvolutional}. However, in the case of a spiking $I(t)$ which transmits relative phase values, we derive a new kernel which allows for exact transformations between these execution domains.

We begin by defining an equivalence between spikes and phases: a decoded relative phase $\theta$ represents a spike at time $t_\text{spike} = (\theta / 2 + 1/2) \cdot T$, where $T$ is the resonant period defined by $2 \pi / \omega$ . If we sample the potential $U(t)$ at discrete steps of $T$, the remaining time before the next sample point is:
\begin{equation}
  \label{eqn_dt}
      \delta t(\theta, k)= (2 \pi / \Im(k)) \cdot (1/2 - \theta / 2)
  \end{equation}
where the period $T = 2\pi/\Im(k)$ is read directly from $k$.
Using the R\&F update equation (Eqn. \ref{eqn_rnf_ssm}), we can calculate the potential contributed by a spike to the neuron's potential at $t_\text{sample}$:
    \begin{equation}
  \label{eqn_dirac_response}
      U'(t_\text{sample}, \theta, k) = e^{k\cdot \delta t({\theta, k})}
  \end{equation}
We may therefore write the following discrete update formula which samples $U$ at steps of $T$ and scales spiking inputs by the weight matrix $B$:
\begin{equation}
\label{eqn_dirac_recurrence}
  U[n+1] = e^{k T} \cdot U[n] + B \cdot U'(T, \theta, k)
\end{equation}
This equation provides the same potential update as provided by the continuous-time neural update (Eqn. \ref{eqn_ssm}) on discrete time-steps. This formulation allows for the system to be updated on discrete time-step, digital neural cores which support complex values (Fig. \ref{fig:equivalence}b).
Furthermore, if we ``unroll'' this update from $U_c[0] = 0$, a causal convolution is produced:
\begin{equation}
    \label{eqn_dirac_conv}
    U[n] = \sum_{j=0}^{n} K[n-j] \cdot B \cdot U'(T, \theta_j, k) = (K * H)[n]
\end{equation}
where $H[n] = B \cdot U'(T, \theta_n, k)$ is defined by the layer's phase inputs and the impulse-response kernel is:
\begin{equation}
    \label{eqn_dirac_kernel}
    K[n] = e^{k \cdot n \cdot T}
\end{equation}
This factorization into $H[n]$  and $K[n]$ separates the computation into an encoding of the input phases and a convolution step which represents the neural updates through time (Fig. \ref{fig:equivalence}c). By taking the FFT of the encoding, we may instead multiply by the kernel K in frequency space and parallelize the computation. After taking the inverse FFT, the relative phase (Eqn.
\ref{eqn_pssm2}) is retrieved by applying Eqn. \ref{eqn_relphase}:
\begin{equation}
\label{eqn_parallel_phase}
  Y_\theta[n] = \theta_{\text{rel}}(U[n], \omega, n \cdot T) = \angle\bigl(U[n] \cdot e^{-i \omega \cdot n \cdot T}\bigr)
\end{equation}
These reformulations define the Phase State Space Model ($\varphi$-SSM), which allows for the advantages of the SSM frameworks to be translated to R\&F networks by providing three equivalent views to execute the network: a continuous differential update which can be implemented by physical systems, a discrete, iterative update which is well-suited to digital hardware, and a kernelized, convolutional view which allows for scale-up and parallelization of training (Table \ref{tab:ssm-costs}). Figure \ref{fig:equivalence}(d-f) demonstrates that our implementations of these methods provide outputs of $U(t)$ which are identical within numerical precision.

\begin{table*}[t]
\centering
\caption{Summary of the computing \& memory costs for the 4 described modes of the $\varphi$-SSM. ``Work'' represents the total number of operations required to calculate the neural states at timestep $L$, and ``Depth'' represents how many of those operations must be calculated in sequence. Continuous/discrete modes require constant memory through time, but calculate sequentially. In contrast, kernelized methods (Toeplitz, FFT) require more memory, but can parallelize the computation to calculate arbitrary neural states quickly.}
\label{tab:ssm-costs}
\begin{minipage}{0.72\linewidth}
\centering
\begin{tabular}{@{}lcccc@{}}
\toprule
& \multicolumn{2}{c}{Compute\textsuperscript{b}} & \multicolumn{2}{c}{Memory} \\
\cmidrule(lr){2-3}\cmidrule(lr){4-5}
Mode & Work & Depth & Infer. & Train \\
\midrule
Continuous ODE      & $\mathcal{O}(s L P\, C B)$ & $\mathcal{O}(L P)$    & $\mathcal{O}(C B)$        & $\mathcal{O}(C B)$\textsuperscript{a} \\
Discrete recurrence & $\mathcal{O}(C L B)$       & $\mathcal{O}(L)$      & $\mathcal{O}(C B)$        & $\mathcal{O}(C L B)$ \\
Toeplitz convolution & $\mathcal{O}(C L^{2} B)$  & $\mathcal{O}(1)$      & $\mathcal{O}(C L^{2}{+}C L B)$ & $\mathcal{O}(C L^{2}{+}C L B)$ \\
FFT convolution     & $\mathcal{O}(C B L \log L)$ & $\mathcal{O}(\log L)$ & $\mathcal{O}(C L B)$      & $\mathcal{O}(C L B)$ \\
\bottomrule
\end{tabular}
\par\smallskip
{\footnotesize
\textsuperscript{a} Via back-solved adjoint equations; naive store-all
backpropagation costs $\mathcal{O}(C B L P)$.\\
\textsuperscript{b} All modes share an input projection $W x(t)$ costing
$\mathcal{O}(C\,C_{\text{in}}\,L\,B)$; the Work column lists only the
temporal-mixing term that distinguishes each mode.}
\end{minipage}%
\hfill
\begin{minipage}{0.25\linewidth}
\centering
\footnotesize
\begin{tabular}{@{}cl@{}}
\toprule
Sym. & Meaning \\
\midrule
$L$             & timesteps computed \\
$C$             & output channels \\
$C_{\text{in}}$ & input features \\
$B$             & batch size \\
$P$             & ODE sub-steps / period \\
$s$             & solver stages \\
\bottomrule
\end{tabular}
\end{minipage}
\end{table*}

\subsection{Connection to HD Computing}

HD computing proposes that high-dimensional vector spaces contain useful geometric characteristics which can be employed via the use of special operators which manipulate points in these HD spaces \cite{kleykoSurveyHyperdimensionalComputing2021}. One HD computing system, the Fourier Holographic Reduced Representation (FHRR), defines operators on the space of points in an HD phase space- a domain identical to the vectors of relative phases which are represented via frequency-locked R\&F neurons which we employ to define the $\varphi$-SSM. The FHRR defines the operations of ``bundling'' (superposition), ``binding'' (rotation), and ``similarity'' (a distance metric) on phases:
  \begin{equation}
  \label{eqn_bundle}
      \texttt{bundle}(\theta_1, \theta_2) = \angle\bigl(e^{i\pi\theta_1} + e^{i\pi\theta_2}\bigr)
  \end{equation}
  \begin{equation}
  \label{eqn_bind}
      \texttt{bind}(\theta_1, \theta_2) = \theta_1 + \theta_2 \mod [-1, 1]
  \end{equation}
  \begin{equation}
  \label{eqn_similarity}
      \texttt{sim}(\boldsymbol{\theta}_1, \boldsymbol{\theta}_2) = \frac{1}{C}\sum_{c=1}^{C} \cos\bigl(\pi(\theta_{1,c} - \theta_{2,c})\bigr)
  \end{equation}
The discrete update of the $\varphi$-SSM can be interpreted in terms of HD operations:
\begin{equation}
  \label{eqn_phase_update}
      Y_\theta[n+1] = \texttt{bundle}\bigl(\texttt{bind}(Y_\theta[n],\, \alpha),\; \theta_{\text{rel}}(H[n], \omega, n \cdot T)\bigr)
\end{equation}
the constant $\alpha = \omega T / \pi$ is the per-step rotation and $H[n] = B \cdot U'(T, \theta_n, k)$ is the encoded input of relative phases. The first argument ``rotates'' the prior state, and the second term contributes the relative phase of the current inputs. Note that when the sampling step equals the resonant period ($T = 2\pi/\omega$), the per-step rotation is a full revolution ($\alpha = 2 \equiv 0$ half-turns), so $\texttt{bind}$ reduces to the identity and the recurrence becomes a pure bundling of the previous state---subject to magnitude decay---with the current input. This is precisely the invariance that frequency-locking provides.

Previously, we demonstrated that these HD operators can themselves be implemented via R\&F neurons. These networks can now be re-interpreted as combining R\&F networks to allow either the innate bundling or binding behavior to be isolated, and similarity may be computed as a simple geometric transform of the interference between multiple R\&F neurons \cite{olin-ammentorpHyperdimensionalComputingProvides2023}.

We propose that this connection allows for cross-pollination of techniques between HD computing and SSMs: the $\varphi$-SSM demonstrates that recurrent HD transformations can be calculated using the ``triple'' view provided by SSM theory, providing new tools for parallel execution and training of HD systems. In parallel, the techniques developed for HD systems for information encoding, memory capacity, and more may be applied to the $\varphi$-SSM \cite{kleykoSurveyHyperdimensionalComputing2021}.

\begin{figure*}[htbp]
    \centering
    \includegraphics[width=1\textwidth]{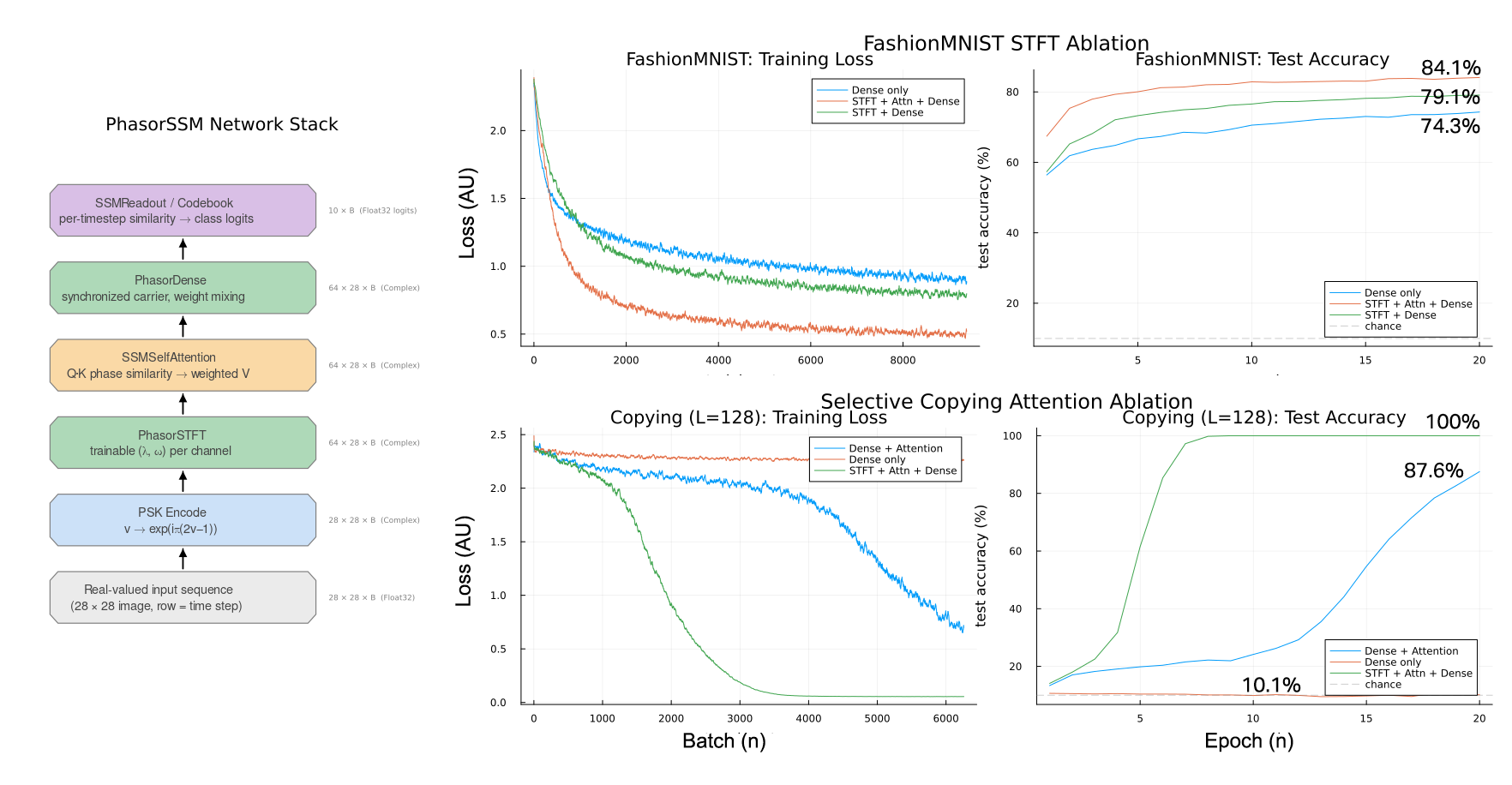}
    \caption{We demonstrate the efficacy of combining the $\varphi$-SSM with other HD computing methods via two tasks: sequential FashionMNIST and a copying task. The addition of the HD computing-based STFT adapter and attentional layer (left) allows both tasks to significantly increase performance (right). In the case of the copying task, the attentional layer is necessary in order to exceed the chance level of performance, and the addition of the STFT adapter allows it to reach full performance on the selective copying task (bottom right).}
    \label{fig:results}
\end{figure*}

\section{Network Demonstration}
We provide a brief demonstration that the $\varphi$-SSM can be integrated with other computational primitives to form a sequence processing network, with all components communicating relative phase values, maintaining compatibility with continuous-time, spiking execution. This network integrates a short-time Fourier transform (STFT), attentional layer, and a similarity-based codebook readout.

\subsection{STFT Adapter}
The $\varphi$-SSM requires the use of frequency-locked neurons to compute with relative phase values which remain invariant through time. While this enables our computational approach developed using constant relative phases, it disallows the ability of the R\&F neuron to resonate with a variety of frequency bands, an ability which can be used to implement useful behaviors such as an STFT \cite{orchardEfficientNeuromorphicSignal2021}.

To resolve this tension, we implement an ``adapter'' layer which performs an STFT via a bank of $C$ multi-compartment R\&F neurons (indexed by $c$). One input compartment has a trainable resonant frequency $\omega_c$ which is driven by a real-valued input signal, giving it an eigenvalue and kernel of:

 \begin{equation}
\label{eqn_stft_kernel}
  k_c = \lambda_c + i\,\omega_c, \qquad K_c[n] = e^{k_c\, n T}.
\end{equation}

In contrast to the $\varphi$-SSM layers, each compartment in this adapter is allowed to have a trainable resonant frequency $\omega_c$. Each input compartment can correspond to one input channel, or several may share the same input to extract multiple frequency components. The potential of this compartment excited by the input is then demodulated by a second compartment's potential produced by an internal, free-running reference oscillator at $\omega_c$. This produces the phase difference between the two compartments, which in turn modulates the final output at the frequency $\omega$ shared with downstream $\varphi$-SSM layers.
\begin{equation}
\label{eqn_stft_out}
  Y_{\theta,c}[n]
   = \theta_{\text{rel}}\!\Big(U_c[n]\,e^{\,i(\omega-\omega_c) n T},\;\omega,\;nT\Big)
   = \theta_{\text{rel}}\big(U_c[n],\;\omega_c,\;nT\big),
\end{equation}
where $U_c[n] = \big(K_c * B X\big)[n]$ is the excited potential of input neuron $c$ at discrete time-step $n$. As the input $X$ is real-valued (a zero-order-hold input rather than a Dirac spike), the per-channel input gain $(e^{k_c T}-1)/k_c$ is absorbed into the projection $B$, leaving the kernel $K_c$ in the form of Eqn.~\ref{eqn_dirac_kernel}.

Thus, by utilizing modulation and demodulation techniques, spiking (or non-spiking) layers which do not inherently compute with phase information can be used to compute and transmit information into $\varphi$-SSM systems. Additionally, a common frequency $\omega$ across many R\&F neurons allows them to communicate using one ``band'' of frequencies using relative phases, but it is possible that more advanced networks may use more than one band, as is observed in biological neural networks \cite{klimeschFrequencyArchitectureBrain2018}. We leave exploration of this possibility to future works.

\subsection{Attention Module}
Attention layers and recurrent layers provide complementary capabilities. While recurrent layers are efficient and can provide a theoretically unlimited history, in practice transformers based on attentional mechanisms can provide higher precision on tasks which require copying of specific information between inputs and outputs \cite{jelassiRepeatMeTransformers}. This module implements self-attention using the $\varphi$-SSM to project inputs into queries, keys, and values, and similarity to compute scores
\cite{olin-ammentorpResidualAttentionalArchitectures2022}.

To produce an attention block, we utilize three $\varphi$-SSM layers ($g_Q,g_K,g_V$) to project a phase input $X_\theta$ of length $L$ into queries, keys, and values ($q_i$, $k_j$, $v_j$, with $i,j \in \{1,\dots,L\}$):
\begin{equation}
\label{eqn_attn_qkv}
  q_i = g_Q(X_\theta)_i,\quad
  k_j = g_K(X_\theta)_j,\quad
  v_j = g_V(X_\theta)_j .
\end{equation}
Attention scores between $q$ and $k$ are the HD similarity (Eqn.~\ref{eqn_similarity}), computable via interference between neurons. These similarities are scaled by a learned factor $\beta$, exponentiated, and normalized by the sequence length $L$ to produce a set of scores $A$:
\begin{equation}
\label{eqn_attn_scores}
  A_{ij} = \frac{1}{L}\,\exp\!\big(\beta\,\texttt{sim}(q_i, k_j)\big).
\end{equation}
By avoiding softmax, the non-local normalization requirement of that function is avoided. These scores can be causally masked, and are then used to selectively bundle (Eqn.~\ref{eqn_bundle}) the values $v_j$ into a combined output:
\begin{equation}
\label{eqn_attn_out}
  O_i = \angle\!\Big(\textstyle\sum_{j} A_{ij}\, e^{\,i\pi v_j}\Big).
\end{equation}

\subsection{Codebook Readout}
Two vectors of relative phase angles can be converted into a single, real-valued score by using the HD similarity operator. Using a fixed codebook of $M$ phase vectors $\{\Phi_m\}$ (random or mutually orthogonal symbols) and computing the similarity of an output vector to these codes, we can predict which class the output corresponds to:

\begin{equation}
\label{eqn_codebook_score}
  s_m = \texttt{sim}\big(Y_\theta,\, \Phi_m\big),
  \qquad
  \hat{y} = \operatorname*{arg\,max}_{m}\, s_m .
\end{equation}

Training a classification network then consists of maximizing the similarity of each output to its correct code, via the loss:

\begin{equation}
\label{eqn_codebook_loss}
  \mathcal{L} = 2\sin^{2}\!\Big(\tfrac{\pi}{4}\big(1 - s_y\big)\Big),
  \qquad s_y = \texttt{sim}\big(Y_\theta,\, \Phi_y\big).
\end{equation}

Again, by avoiding softmax non-local normalizations are avoided, and calculations are expressed using HD operators which can potentially be implemented on novel analog or digital hardware platforms.

\subsection{Results \& Discussion}
A synthetic data-copying task and sequential FashionMNIST were used to benchmark networks integrating these components. Each hidden layer implemented via $\varphi$-SSM utilized
64 R\&F neurons, and networks were trained for 20 epochs. In both cases, the STFT adapter and attention module improved performance as demonstrated through an ablation test. In the copying task, where the network must ``remember'' one input in a long series, the attention module was required in order to exceed chance levels of performance, improving to a perfect output with the addition of the STFT module (Fig. \ref{fig:results}).

While these results were demonstrated using training achieved via the parallel, kernelized execution method, we posit that it may be possible to extend training methods to the discrete and continuous scenarios via defining adjoint equations that could be used to update eligibility traces for individual neurons \cite{rackauckasUniversalDifferentialEquations2021, bellecEligibilityTracesProvide2019}. Alternatively, exploring alternative connectivity patterns between layers of an $\varphi$-SSM could implement the ``feedback'' connections used for equilibrium propagation, allowing for neural oscillations to innately encode local gradients \cite{laborieuxHolomorphicEquilibriumPropagation2022}.

\section{Conclusion}

By utilizing layers of R\&F neurons which share a common resonant frequency, we defined a recurrent system which transforms vectors of relative phase values which can remain invariant through time. Applying SSM theory to this system, we derived the $\varphi$-SSM: a system which provides multiple, equivalent execution methods which support both efficient inference and highly parallel training. Furthermore, the additional equivalence of representations and operations between the $\varphi$-SSM and HD computing allows for further cross-pollination between two productive areas of research. We demonstrated that including HD methods along with a $\varphi$-SSM network is both possible and beneficial to performance on two simple benchmarks. We believe this encouraging preliminary result motivates further exploration of these methods applied to more complex architectures and tasks, as well as providing new possibilities for physical computing systems based on R\&F neurons to be efficiently simulated and trained \textit{in-silico}.

\begin{acks}
This work was supported by DOE ASCR BIA: A Co-Design Methodology to Transform Materials and Computer Architecture Research for Energy Efficiency. This material is based upon work supported by the U.S. Department of Energy, Office of Science, under contract number DE-AC02-06CH11357.

The submitted manuscript has been created by UChicago Argonne, LLC, Operator of Argonne National Laboratory ``Argonne''). Argonne, a U.S. Department of Energy Office of Science laboratory, is operated under Contract No. DE-AC02-06CH11357. The U.S. Government retains for itself, and others acting on its behalf, a paid-up nonexclusive, irrevocable worldwide license in said article to reproduce, prepare derivative works, distribute copies to the public, and perform publicly and display publicly, by or on behalf of the Government. The Department of Energy will provide public access to these results of federally sponsored research in accordance with the DOE Public Access Plan. \hyperlink{http://energy.gov/downloads/doe-public-access-plan}{http://energy.gov/downloads/doe-public-access-plan}

AI was applied in this work to generate proofs and code, create figures, collect relevant literature, and review drafts for typos and other grammatical errors. No AI-generated text has been used in this manuscript.

Finally, we thank the reviewers for their time and suggestions to improve the final version of this work.
\end{acks}

\bibliographystyle{ACM-Reference-Format}
\bibliography{references}

\appendix

\section{Online Resources}

The software package ``PhasorNetworks.jl'' implementing the $\varphi$-SSM and our experiments can be found online at \url{https://github.com/wilkieolin/PhasorNetworks.jl} .\

\end{document}